\pdfoutput=1

\documentclass[11pt]{article}

\usepackage[]{acl2023} 

\usepackage{times}
\usepackage{latexsym}

\usepackage{amsmath,amsfonts,bm}

\def\eqref#1{equation~\ref{#1}}

\def\1{\bm{1}}

\DeclareMathAlphabet{\mathsfit}{\encodingdefault}{\sfdefault}{m}{sl}
\SetMathAlphabet{\mathsfit}{bold}{\encodingdefault}{\sfdefault}{bx}{n}

\usepackage{hyperref}

\usepackage{url}
\usepackage{graphicx}
\usepackage{wrapfig}
\usepackage{multirow}
\usepackage{float,booktabs}
\usepackage{amssymb}
\usepackage{pifont}
\newcommand{\cmark}{\ding{51}}%
\newcommand{\xmark}{\ding{55}}%

\usepackage[T1]{fontenc}

\usepackage[utf8]{inputenc}

\usepackage{microtype}

\usepackage{inconsolata}

\NewDocumentCommand{\zhenhailong}{ mO{} }{\textcolor{cyan}{\textsuperscript{\textit{zhenghailong}}\textsf{\textbf{\small[#1]}}}}

\usepackage{xparse}
\NewDocumentCommand{\heng}
{ mO{} }{\textcolor{red}{\textsuperscript{\textit{Heng}}\textsf{\textbf{\small[#1]}}}}

\newcommand{\ours}{\text{Zemi}}
\newcommand{\zb}{\text{Zemi$_\text{BASE}$}}
\newcommand{\zl}{\text{Zemi$_\text{LARGE}$}}
\newcommand{\noaug}{\text{No Aug}}
\newcommand{\noaugb}{\text{No Aug$_\text{BASE}$}}

\newcommand{\concat}{\text{Concat}}

\newcommand{\fid}{\text{FiD}}
\newcommand{\fusion}{\text{retrieval-augmentation fusion}}

\title{\ours{}: Learning Zero-Shot Semi-Parametric Language Models from Multiple Tasks}

\author{
Zhenhailong Wang\thanks{\;\;Work was done when interning at Tencent AI Lab.}\\ UIUC\\ \texttt{\small{wangz3@illinois.edu}}\\
\And
Xiaoman Pan\\Tencent AI Lab\\\texttt{\small{xiaomanpan@global.tencent.com}}
\And
Dian Yu\\Tencent AI Lab\\\texttt{\small{yudian@global.tencent.com}}
\AND
Dong Yu\\Tencent AI Lab\\\texttt{\small{dyu@global.tencent.com}}
\And
Jianshu Chen\\Tencent AI Lab\\\texttt{\small{jianshuchen@global.tencent.com}}
\And
Heng Ji\\UIUC\\\texttt{\small{hengji@illinois.edu}}
}

\begin{document}

\maketitle

\begin{abstract}
Although large language models have exhibited impressive zero-shot ability, the huge model size generally incurs high cost. Recently, semi-parametric language models, which augment a smaller language model with retrieved related background knowledge,
alleviate the need for storing everything into the model parameters.
Although existing semi-parametric language models have demonstrated promising \textit{language modeling} capabilities, it remains unclear whether they can exhibit competitive \textit{zero-shot} abilities as their fully-parametric counterparts.
In this work, we introduce \textbf{\ours{}}, 
a semi-parametric language model for zero-shot task generalization. To our best knowledge, this is
\textbf{the first semi-parametric language model that can demonstrate strong zero-shot performance on a wide range of held-out unseen tasks.} We train \ours{} with semi-parametric multitask training, which shows significant improvement compared with the parametric multitask training as proposed by T0~\citep{t0}. Specifically, during both training and inference, \ours{} is equipped with a retrieval system based on the unlabeled pretraining corpus of our backbone model.
To address the unique challenges from large-scale retrieval, we further propose a novel \textbf{\fusion{}} module that can effectively incorporate noisy retrieved documents.
Finally, we show detailed analysis and ablation studies on the key ingredients towards building effective zero-shot semi-parametric language models.
Notably, our proposed \zl{} model outperforms T0-3B by 16\% across seven diverse evaluation tasks while being 3.8x smaller in scale.\footnote{Code and data are available for research purpose at \url{https://github.com/MikeWangWZHL/Zemi}.}

\end{abstract}

\begin{figure*}[t]
    \centering
    \includegraphics[width=0.95\textwidth]{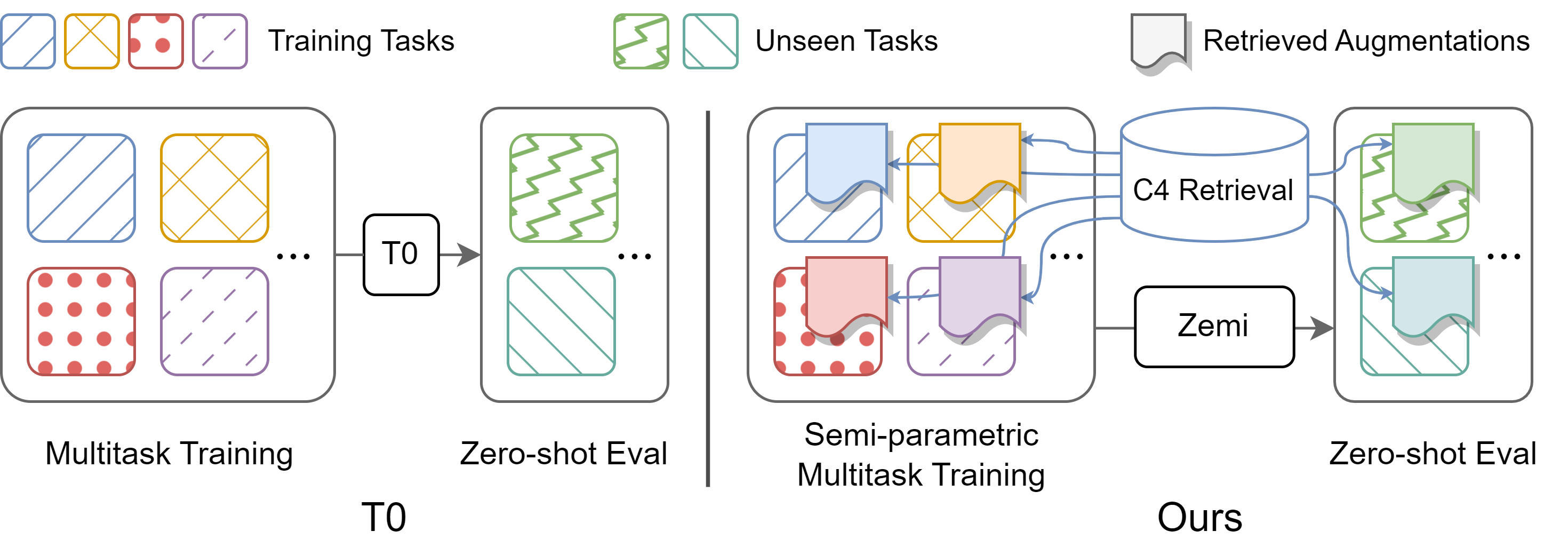}
    \vspace{-5pt}
    \caption{Overview of the semi-parametric multitask prompted training. Each training and evaluation instance is formatted with unified text-to-text prompt templates~\citep{t0, promptsource}. In this work, we further augment the prompted instances with retrieved passages from a large-scale task-agnostic corpus, C4~\citep{t0}, which is the same unlabeled pretraining corpus used in T5~\citep{t5} and T0~\citep{t0}. An example of the prompted input and the retrieved documents can be found in Figure~\ref{fig:architecture}.}
    \label{fig:overview}
    \vspace{-5pt}
\end{figure*}

\section{Introduction}

Achieving strong generalization ability on unseen tasks while maintaining a reasonably small parameter size is a long-lasting challenge for natural language processing (NLP) models. Although large language models~\citep{gpt3,jurassic-1,gopher,megatron-turing,chinchilla,OPT,instructGPT,PaLM} have shown impressive zero-shot ability on various NLP tasks, the huge model size generally incurs high cost. Alternatively, instead of stuffing everything in the model parameters, recent work on semi-parametric language models~\citep{cache_lm, knn-lm, spalm, retro, memory_aug_lm} demonstrated competitive \textit{language modeling} performance compared with much larger fully-parametric language models. The intuition is to use a relatively small language model as a reasoning module and augment it with a retriever to retrieve related background knowledge,
which effectively alleviates the need for increasing the model capacity to align with the growing data size.

However, what really makes large language models the focus of attention in the past two years is their strong zero-shot in-context learning abilities.
Unfortunately, it is still unclear whether semi-parametric language models can exhibit similar \textit{zero-shot} ability on unseen tasks as their fully-parametric counterparts such as T0~\citep{t0} and GPT-3~\citep{gpt3}. Moreover, improvements in language modeling metrics such as perplexity may not guarantee better performance on downstream tasks especially in low-shot settings~\citep{wei2022emergent}. 
Thus, in this work, we aim to investigate this unexplored research question, \textit{can semi-parametric language models exhibit strong zero-shot generalization abilities on various downstream tasks?}

To this end, we introduce \textbf{\ours{}}, a \underline{z}ero-shot s\underline{emi}-parametric language model. To the best of our knowledge, this is the first semi-parametric language model that shows strong zero-shot performance on a wide range of downstream tasks. In order to effectively train \ours{}, we propose to extend the multitask prompted training~\citep{t0} into semi-parametric settings (Section \ref{sec:training_paradigm}). Specifically, during both the training and the inference stage, we augment the prompted instances with retrieved plain text documents. To cover a wider range of unseen tasks, instead of retrieving from specific corpora for certain tasks, such as exploiting Wikipedia for open-domain question answering~\citep{ORQA,DPR,FiD}, we retrieve documents from a large-scale task-agnostic corpus, C4~\citep{t5} (Section \ref{sec:c4_retrieval}). Notably, C4 is the unlabeled pre-training corpus of our backbone model~\citep{t5}, which means that every document is seen by the model and we do not require any annotated or curated resources. This guarantees fair comparison with the parametric counterpart T0~\citep{t0}. 

In our preliminary experiments, we find that existing methods~\citep{FiD,gpt3} for incorporating retrieved text cannot effectively handle the noise inevitably introduced by retrieving from large-scale corpora. 
To address this challenge, we propose a novel \textbf{retrieval-augmentation fusion} module that can selectively ignore noisy retrieved text.
Specifically, we introduce a light-weight \textit{perceiver resampler} and a \textit{gated cross-attention} layer~\citep{flamingo} to enforce the model to attend to salient information of each augmentation and gate out noisy ones (Section~\ref{sec:augmentation_fusion}).

We train \ours{} on eight multiple-choice question answering (QA) tasks (4.5x fewer than T0) and evaluate on a diverse set of seven unseen tasks from five categories (Section~\ref{sec:datasets}). 
In order to investigate the impact of the retrieval-augmentation, we favor knowledge-intensive tasks over extractive tasks.

Experimental results show that \ours{} outperforms both parametric and semi-parametric baselines. Notably, \zl{} outperforms T0-3B by 16\% across seven evaluation tasks while being 3.8x smaller in scale (Section~\ref{sec:compare_to_sota}). We further conduct extensive analysis on \textit{why \ours{} works}. We show that the source of the improvements comes from the interplay of our proposed \fusion{} architecture along with the semi-parametric multitask training paradigm. Finally, we perform in-depth ablation studies on all aspects of our model design including the gated mechanism.

To sum up, the main contributions of this paper are threefold:
\begin{itemize}
    \vspace{-4pt}
    \item We introduce \ours{}, which is to our knowledge the first semi-parametric model that demonstrates strong zero-shot task generalization ability. 
    \vspace{-4pt}
    \item We propose a novel retrieval-augmentation fusion module which can effectively handle multiple potentially noisy retrieved documents and is essential towards the effectiveness of semi-parametric multitask training.
    \vspace{-4pt}
    \item We show detailed analysis and ablation studies on \textit{why \ours{} works} which shed light on future work for developing large-scale universal semi-parametric language models with strong zero-shot ability.  
\end{itemize}

\section{Method}

\subsection{Semi-parametric multitask training}
\label{sec:training_paradigm}
In this section, we introduce how we extend the multitask training paradigm to semi-parametric language models. We follow the overall text-to-text framework proposed by the previous parametric multitask prompted training~\citep{t0} where each input-output pair of a certain task is converted into a prompted text input and a generated text output via human-written templates~\citep{promptsource}.\footnote{\href{https://github.com/bigscience-workshop/promptsource}{https://github.com/bigscience-workshop/promptsource}.}
For \ours{}, as illustrated in Figure~\ref{fig:overview}, during both training and inference, we augment \ours{} with a retrieval system. Instead of using specific corpora for different tasks, such as Wikipedia for open-domain question answering~\citep{DrQA,DPR,FiD} and textbooks for science question answering~\citep{OpenBookQA}, we retrieve texts from a large-scale task-agnostic corpus, \textbf{C4}~\citep{t5} (Section~\ref{sec:c4_retrieval}). Retrieving from a larger corpus brings wider coverage but also more noisy augmentations. To address this problem we further propose a novel semi-parametric architecture for \ours{} that specializes in handling a large number of potentially noisy augmentations (Section~\ref{sec:augmentation_fusion}). After semi-parametric multitask training, we perform zero-shot evaluation on seven diverse held-out unseen tasks (Section~\ref{sec:experiments}).

\subsection{C4 retrieval}
\label{sec:c4_retrieval}

To build a universal semi-parametric language model that can generalize to various types of NLP tasks, we retrieve documents from Colossal Clean
Crawled Corpus (C4)~\citep{t5}. 
Notably, C4 is the unlabeled pre-training corpus of our backbone model T5~\citep{t5}, which guarantees fair comparison with non-retrieval methods in our zero-shot evaluation settings.
The C4 corpus (750GB in size) contains more than 364 million documents. Performing dense retrieval~\citep{DPR} on such a wide-coverage corpus is very expensive. Thus, for efficiency consideration, we perform document-level indexing and retrieval based on BM25~\citep{bm25} with ElasticSearch~\citep{elasticsearch} and Huggingface Datasets~\citep{hfdatasets}. Despite its simplicity, recent work~\citep{wang2022training} has demonstrated the effectiveness of using BM25 for retrieving clean training data as augmentations. To further improve the retrieval efficiency, we use 5\% of the entire C4 corpus, which is still 3x larger than the Wikipedia corpus~\citep{wikidump}, as our retrieval corpus in our experiments. For each query, we truncate the query length at 20 tokens and truncate each retrieved document at 256 tokens. See details on the query fields for each dataset in Appendix~\ref{app:query_key}.

\begin{figure*}[thb]
    \centering
    \includegraphics[width=\textwidth]{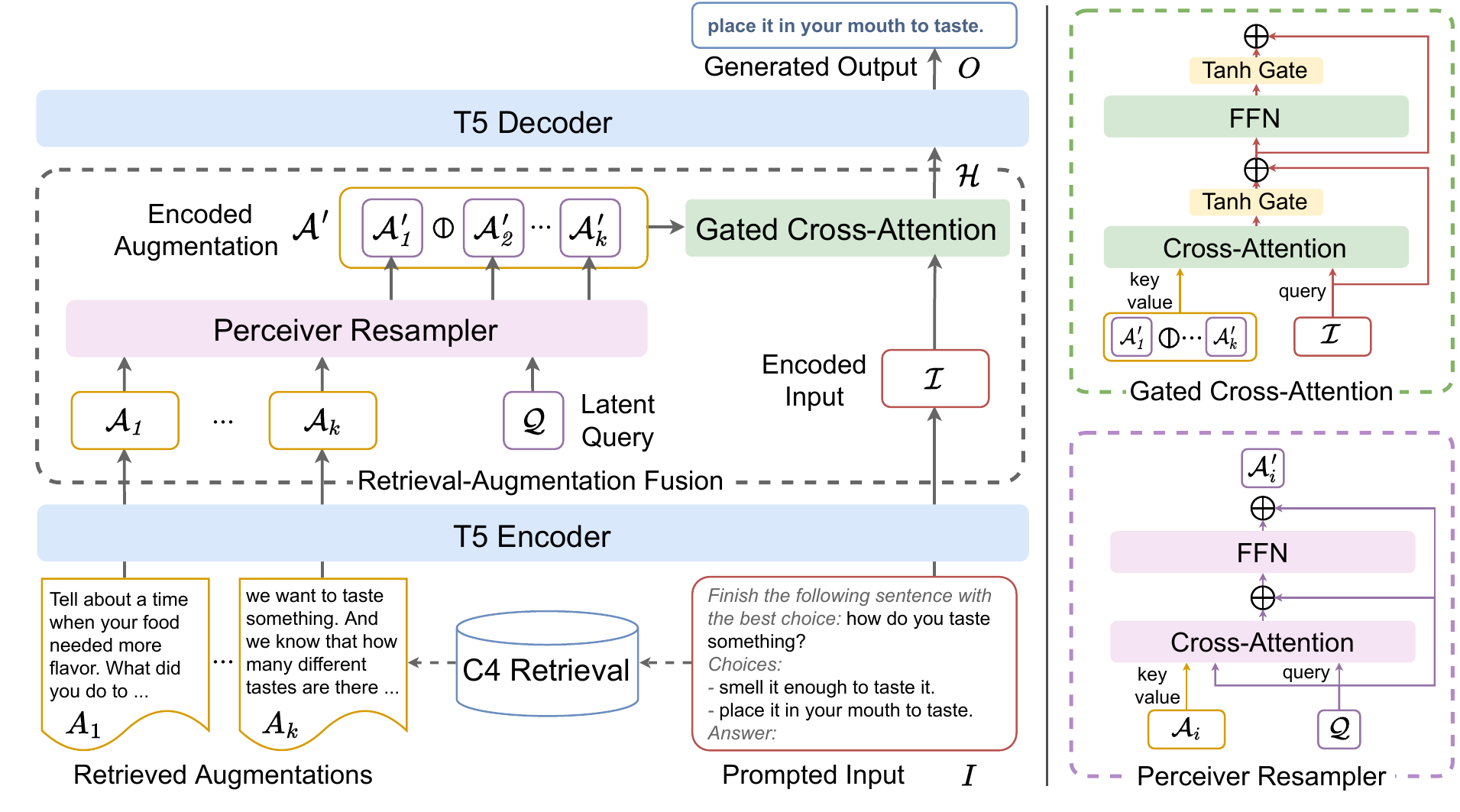}
    \vspace{-5pt}
    \caption{\ours{} model architecture with an example of a prompted input and a generated output from the Piqa~\citep{piqa} task. The \textit{italic text} in the prompted input $I$ indicates the prompt template. $A_1$ and $A_k$ shows two examples of the corresponding retrieved augmentations (documents) from the C4 corpus. To incorporate the potentially noisy retrieved augmentations, we introduce a light-weight \fusion{} module that contains two major components, a single layer perceiver resampler and a single layer gated cross-attention (detailed on the right).
    }
    \vspace{-10pt}
    \label{fig:architecture}
\end{figure*}

\subsection{\ours{} model architecture}
\label{sec:augmentation_fusion}

One major challenge of retrieving from a large-scale task-agnostic corpus is that the retrieved augmentations (documents) can be noisy and inaccurately ranked. Examples of good and noisy retrieved documents can be found in Appendix~\ref{sec:retrieval_examples}. To address this problem, intuitively, we want the model to have the following two properties: (1) be able to simultaneously pay attention to multiple retrieved augmentations instead of only the top-1 document. (2) be able to identify salient information from the retrieved augmentations and selectively ignore uninformative ones.

To this end, we propose the \ours{} architecture, a semi-parametric language model capable of selectively incorporating multiple potentially noisy retrieved augmentations.
The main idea is to jointly train a light-weight \textbf{\fusion{}} module between the encoder and decoder, which contains two major components, a \textit{perceiver resampler} and a \textit{gated cross-attention}, which are inspired by recent work on vision-language fusion~\citep{flamingo}. 

Figure~\ref{fig:architecture} shows an illustration of the \ours{} model architecture. 
We consider a prompted text input $I$ and a few retrieved textual augmentations $A_1, A_2,...,A_k$. Let $l_I$, $l^i_A$ be the length of the prompted input and the ith augmentation. Let $d$ be the hidden dimension of our backbone model. We first independently encode $I$ and $A_1, A_2,...,A_k$ with a shared T5~\citep{t5} encoder $\mathrm{Enc}$.
We then feed the latent representation of the augmentations $\mathcal{A}_{1},\mathcal{A}_{2},...,\mathcal{A}_{k}$ through the perceiver resampler.
\begin{align} 
    \mathcal{I} &= \mathrm{Enc}(I)\\
    \mathcal{A}_i &= \mathrm{Enc}(A_i)\\
    \mathcal{A}'_i &= \mathrm{PerceiverResampler}(\mathcal{A}_i,\mathcal{Q})
\end{align}
where $\text{$\forall$ $ i\in\{1,..,k\}$}$, $\mathcal{I} \in R^{l_I \times d}$, $\mathcal{A}_{i} \in \mathcal{R}^{l^i_A \times d}$ and $\mathcal{A}'_i \in \mathcal{R}^{l_Q \times d}$.

As shown on the bottom right of Figure~\ref{fig:architecture}, the perceiver resampler is a variant of Perceiver IO~\citep{perceiver_IO}, where a cross-attention is performed between the variable-length latent representation of an augmentation $\mathcal{A}_{i}$ and a fixed-length learnable latent query vector $\mathcal{Q}$. 
Let $l_Q$ be the predefined length of the latent query, which is typically smaller than the original length of an augmentation $l^i_{A}$. The output of the perceiver resampler is a compressed fixed-length latent representation of each augmentation.
This resampling mechanism not only allows the model to include \textit{longer and a larger number of augmentations} but also encourages the model to select \textit{salient information} from the original augmentations.
After the resampler, we concatenate the encoded augmentations $\mathbf{\mathcal{A'}} = [\mathcal{A}'_i,..,\mathcal{A}'_k]$ and perform gated cross-attention with the encoded prompted input $\mathcal{I}$. As shown in the top right of Figure~\ref{fig:architecture}, the gated cross-attention layer contains two Tanh gates controlling the information flow from the cross-attention layer and the feed-forward layer before the addition with the skip connections, i.e., the original encoded input $\mathcal{I}$.
Finally, the hidden states from the gated cross-attention module $\mathcal{H}$ is fed into the T5 decoder $\mathrm{Dec}$ to generate the output sequence $O$.
\begin{align}
    \mathcal{H} &=\mathrm{GatedCrossAttn}([\mathcal{A}'_i,..,\mathcal{A}'_k], \mathcal{I})\\
    O &= \mathrm{Dec}(\mathcal{H})
\end{align}
where $[\mathcal{A}'_i,..,\mathcal{A}'_k] \in R^{(k \times l_Q) \times d}$ and $\mathcal{H} \in \mathcal{R}^{l_I \times d}$.

Following \cite{flamingo}, we initialize the parameter of the Tanh gate to be 0, allowing the forward pass of the prompted input through the pre-trained T5 encoder-decoder to be intact at the beginning of the training process. 
With the gated mechanism, the model can learn to \textit{gate out noisy augmentations} and rely more on the skip connections during semi-parametric multitask training.

\section{Experiments}
\label{sec:experiments}

\vspace{-5pt}
\subsection{Experimental setup}
\label{sec:datasets}

\begin{table*}[t]
\footnotesize
\centering
\setlength\tabcolsep{3pt}
\begin{tabular}{c | c  c  c | l  l  l  l  l  l  l | l l}
\toprule

\multirow{2}{*}{\textbf{Method}} &
\multirow{1}{*}{\textbf{semi-}} &
\multirow{1}{*}{\textbf{\# train}} &
\multirow{2}{*}{\textbf{\# param}} &
\multicolumn{7}{c|}{\textbf{Tasks}} &
\multirow{2}{*}{Avg$_5$} & 
\multirow{2}{*}{Avg$_7$} 
\\

& \textbf{param} & \textbf{tasks} &  & 
OBQA & Piqa & RT & CB & COPA & WiC & HSwag & 
\\
\midrule

BART0     & No & 36 & 0.4B      & 34.4 & 36.1      & - & 39.6       & -    & 46.7        & 39.4 & 39.3 & - \\
T0-3B                        & No & 36 & 3B        & 42.8 & 59.3      & 73.6$^\star$ & 45.5  & 75.9   & 50.0        & 27.3  & 45.0 & 53.5\\
T0-11B                     & No  & 36 & 11B       & 59.1 & 72.5      & 81.8$^\star$ & 70.1  & 91.5   & 55.2        & 33.5 & 58.1 & 66.3 \\
\midrule
ReCross  & Yes &  36 & 0.4B     & 39.6 & 41.4      & - & 44.8       & -    & 50.6        & 47.3  & 44.7 & - \\
\textbf{\zb{} (ours)}          & Yes & 8  & 0.2B     & 35.6 & 59.2 & 68.6 & 50.1 & 63.6 & 49.6 & 29.7 & 44.8 & 50.9 \\
\textbf{\zl{} (ours)}         & Yes & 8  & 0.8B     & 51.5 & 67.9 & 84.1 & 62.1 & 84.5 & 50.4 & 35.8 & 53.5 & 62.3 \\
\midrule
GPT-3                 & No & -  & 175B      & 57.6 & 81.0$^\star$  & 59.7 & 46.4       & 91.0   & \;49.4$^\dagger$   & 78.9 & 62.7 & 66.3\\


\bottomrule
\end{tabular}
\caption{Comparison to both parametric (\textit{BART0, T0, GPT-3}) and semi-parametric (\textit{ReCross}) state-of-the-art. \textit{\zl{}} significantly outperforms T0-3B while being 3.8x smaller  in scale. \textit{\zb{}} slightly outperforms \textit{ReCross} while being 1.7x smaller. 
Note that Avg$_7$ indicates averaged performance across all seven tasks. Avg$_5$ indicates averaged performance on five tasks excluding \textit{RT} and \textit{COPA} due to unreported baseline results. $^\star$ indicates the task is seen during training. $^\dagger$ indicates few-shot results with 32 examples.}
\label{table:comparison_SOTA}
\vspace{-5pt}
\end{table*}

Following \cite{t0} we partition various types of NLP tasks into two groups, training tasks and held-out unseen tasks. In this work, we are particularly interested in investigating the impact of the retrieval augmentation
Thus, when choosing the training and evaluation tasks, we favor knowledge-intensive tasks over extractive tasks such as summarization,
where most knowledge for solving the problem is already self-contained in the input. Furthermore, we avoid including large datasets, such as DBPedia~\citep{dbpedia} (630K instances) and QQP~\citep{qqp} (400K instances), due to limited computational resources.

\paragraph{Training Tasks} We use a subset of T0's~\citep{t0} training mixture for our semi-parametric multitask prompted training. Specifically, our training mixture contains \textit{eight} multiple-choice QA datasets, including, \textbf{CommonsenseQA}~\citep{cos_e}, \textbf{CosmosQA}~\citep{cosmosqa}, \textbf{DREAM}~\citep{dream}, \textbf{QASC}~\citep{qasc}, \textbf{QUARTZ}~\citep{quartz}, \textbf{SciQ}~\citep{SciQ}, \textbf{Social IQa}~\citep{socialiqa}, and \textbf{WIQA}~\citep{wiqa}. We choose the subset in multiple-choice QA tasks because they are diverse in domains and overall task formats. Ablation studies on including more types of training tasks can be found in Section~\ref{sec:ablation}.

\paragraph{Evaluation Tasks} For evaluation tasks, we consider \textit{seven} datasets from \textit{five} diverse categories following the task taxonomy of T0, including, two sentence completion tasks, \textbf{COPA}~\citep{COPA} and \textbf{HellaSwag} (HSwag)~\citep{hellaswag}, two multiple-choice QA tasks, \textbf{OpenbookQA} (OBQA)~\citep{OpenBookQA} and \textbf{Piqa}~\citep{piqa}, one word sense disambiguation task, \textbf{WiC}~\citep{wic}, one sentiment task, \textbf{Rotten Tomatoes} (RT)~\citep{rotten_tomatoes}, and one natural language inference task, \textbf{CB}~\citep{CB}. 
All scores are reported on the validation set of each dataset. The detailed prompt templates used for training and evaluation can be found in Appendix~\ref{app:templates}.

\paragraph{Prompts} We use \textit{PromptSource}~\citep{promptsource} with \textit{Huggingface Datasets}~\citep{hfdatasets} to construct prompted inputs for each training and evaluation instance. During training, we randomly select two templates for each dataset. During evaluation, we follow the exact evaluation procedure as in T0~\citep{t0} and report the mean accuracy across all available templates. All scores are reported on the validation set of each dataset. The detailed templates used for training and evaluation can be found in Appendix~\ref{app:templates}.

\paragraph{Model} We consider two variants of \ours{} with a different pre-trained backbone, i.e., T5-base and T5-large~\citep{t5}. Following T0~\citep{t0}, we use the language modeling adapted\footnote{\href{https://huggingface.co/google/t5-base-lm-adapt}{https://huggingface.co/google/t5-base-lm-adapt}.} checkpoint, which is trained for an additional 100k steps on a language modeling objective.
By default, we use five retrieved passages as augmentations for each instance. More implementation details can be found in Appendix~\ref{app:implementation_details}

\subsection{Main results}
\label{sec:compare_to_sota}
We aim to explore whether \ours{} can exhibit competitive zero-shot performance against larger state-of-the-art language models. We compare \ours{} with both parametric (\textit{T0}~\citep{t0}, \textit{BART0}~\citep{Lin2022UnsupervisedCG}, \textit{GPT-3}~\citep{gpt3}) and semi-parametric (\textit{ReCross}~\citep{Lin2022UnsupervisedCG}) models on seven zero-shot tasks. Table~\ref{table:comparison_SOTA} shows the mean zero-shot accuracy across all templates for each task. The last two columns of Table~\ref{table:comparison_SOTA} show the averaged performance across different sets of tasks, where Avg$_7$ is averaged across all seven tasks, and Avg$_5$ considers five tasks excluding RT and COPA due to their unavailable baseline results. 

For \textit{BART0}, \textit{ReCross}, and \textit{GPT-3}, we copy the reported scores directly from their original papers. For the missing score of RT on \textit{GPT-3}, we run the original text completion API~\footnote{For consistency with other results, we report the RT result from the original ``davinci'' model.} to get the generated outputs which is then mapped to the most similar answer choice using SentenceBert~\citep{sentence-bert}. For \textit{T0} models, there are some tasks such as OBQA and Piqa that are not evaluated in the original paper~\citep{t0}, and some tasks such as CB and WiC are evaluated with slightly different templates. Thus, for fair comparison, we re-evaluate all seven tasks on T0-3B and T0-11B using the official implementation and checkpoints\footnote{\href{https://github.com/bigscience-workshop/t-zero}{https://github.com/bigscience-workshop/t-zero}.} with the exact same set of templates as our model. See details on the templates used for each task in Appendix~\ref{app:templates}.

\paragraph{Result} Table~\ref{table:comparison_SOTA} shows that \zb{} outperforms previous retrieval-based method, ReCross, on the average of five tasks (Avg$_5$) while being 2x smaller in scale. Notably, \textbf{\zl{}, significantly outperforms T0-3B} on seven evaluation tasks (Avg${_7}$) by 16\% with \textbf{3.8x fewer parameters}. This shows that \ours{} scales up well with larger backbone models. We also observe that although trained with 4.5x fewer training tasks (8 v.s. 36), \ours{} effectively achieves state-of-the-art zero-shot performance. In Section~\ref{sec:ablation}, we show that adding more tasks into multitask training does not necessarily improve the performance. And the training mixture with multiple-choice QA tasks seems to be highly effective in generalizing to various kinds of unseen tasks.

\begin{table*}[t]
\centering
\setlength\tabcolsep{3pt}
\begin{tabular}{c c | c  c  c  c  c  c  c |  c }
\toprule

\multirow{2}{*}{\textbf{Method}} &
\multirow{2}{*}{\textbf{\# Param}} &
\multicolumn{7}{c|}{\textbf{Tasks}} &
\multirow{2}{*}{Avg} \\
& & OBQA & Piqa & RT & CB & COPA & WiC & HSwag & 
\\
\midrule
\textbf{\noaug}  & 0.8B & 50.5 & 65.5 & 82.2 & 52.4 & 80.0 & 50.2 & 34.1 & 59.3 \\

\midrule

\textbf{\concat}  & 0.8B & 48.8 & 65.9 & 74.9 & 44.6 & 82.7 & 50.0 & 30.5 & 56.8\\
\textbf{\fid}   & 0.8B & 51.0 & 66.7 & 67.1 & 60.7 & \textbf{86.3} & 50.2 & 32.9 & 59.3  \\
\textbf{\zl{} (Ours)} & 0.8B & \textbf{51.5} & \textbf{67.9} & \textbf{84.1} & \textbf{62.1} & 84.5 & \textbf{50.4} & \textbf{35.8} & \textbf{62.3} \\

\bottomrule
\end{tabular}
\caption{Comparison to parametric multitask trained baseline (\noaug) and alternative augmentation fusion methods (\concat, \fid) with an identical backbone model, T5-large. \# Param indicates the model size. }
\label{table:main_result}

\end{table*}

\subsection{Analysis: semi-parametric v.s. parametric}
\label{sec:semi-vs-vanilla}
In order to further analyse the source of the strong performance of \textit{\zl{}}, we compare \textit{\textbf{\zl{}}} with a baseline (\textit{\textbf{No Aug}}) trained with parametric multitask training on the same set of training tasks and with the same backbone model, T5-Large~\cite{t5}. To show the impact of our newly proposed \fusion{} module, we further compare \textit{\zl{}} against two semi-parametric baselines with a different fusion method for incorporating the retrieved augmentations (\textit{\textbf{Concat}} and \textit{\textbf{FiD}}). 
In Table~\ref{table:main_result}, we show that the source of benefit comes from \textbf{the interplay of the proposed \fusion{} and the semi-parametric multitask training}. 

Specifically, for \textit{Concat}, we directly concatenate all retrieved augmentations with the prompted input text. The concatenated input is then truncated to the maximum acceptable length of 1024 tokens and fed to our backbone model. For \textit{FiD}, we implement the model following \cite{FiD} where we first independently encode each pair of retrieved augmentation and the prompted input text. Then we concatenate the encoder outputs and feed them to the decoder. Note that we keep everything else identical except the \fusion{} module for \textit{\zl{}}, \textit{Concat} and \textit{FiD}.

\begin{figure}[t]
    \centering
    \includegraphics[width=0.5\textwidth]{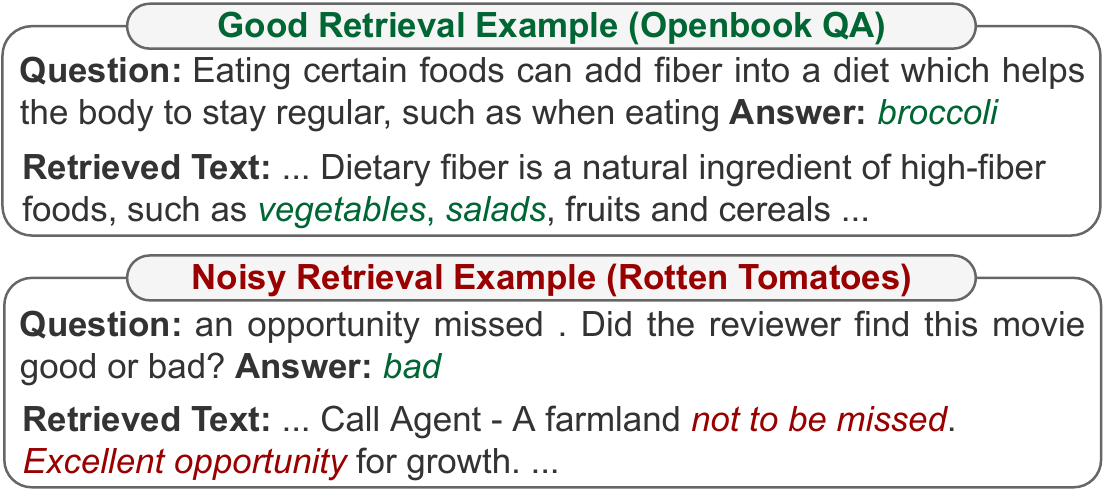}
    \caption{Example of good and noisy retrieved augmentations. See Appendix~\ref{sec:retrieval_examples} for more examples.}
    \label{fig:visualization_short}
\end{figure}

\paragraph{\ours{} architecture improves zero-shot task generalization.} 
In Table~\ref{table:main_result}, we first notice that the semi-parametric setting in itself does not necessarily bring consistent positive gains compared with the \textit{\noaug{}} baseline, as shown in the results of \textit{\concat{}} and \textit{\fid{}}. This can be explained by the fact that the retrieved documents are not always highly correlates with the task of interest, as shown in the examples in Figure~\ref{fig:visualization_short}. The fact that \textit{\fid{}} performs better than \textit{\concat{}} further verifies this hypothesis, since \textit{\fid{}} preserves more input text information in the encoding step and only do fusion with all the retrieved augmentations in the decoder, whereas \textit{\concat{}} perform unified self-attention on all augmentations concatenated directly to the input.

On the other hand, with the proposed \fusion{} module that contains the explicit resampling and gating mechanism, \textit{\zl{}} was able to achieve the best performance on six out of seven tasks, and brings a overall gain of \textbf{+5\%} against the \textit{\noaug{}} baseline. This result shows that the \fusion{} module in \textit{\ours{}} can effectively enable the model to leverage potentially noisy retrieved augmentations during semi-parametric multitask training, which brings significant improvement in zero-shot task generalization. In abaltion study~\ref{sec:ablation}, we further verify that the gated cross-attention is an important factor contributing to the effectiveness of the \ours{} architecture.

\subsection{Analysis: ablation studies}
\label{sec:ablation}

\begin{table*}[!hbt]
\centering
\setlength\tabcolsep{3pt}
\begin{tabular}{l l | l l | c  c  c  c  c  c  c | c }
\toprule
&
\multirow{1}{*}{\textbf{Ablated}} &
\multirow{1}{*}{\textbf{\ours{}}} &
\multirow{1}{*}{\textbf{Changed}} &
\multicolumn{7}{c|}{\textbf{Tasks}} &
\multirow{2}{*}{Avg}\\

& \textbf{setting} & \textbf{value} & \textbf{value} &
OBQA & Piqa & RT & CB & COPA & WiC & HSwag &
\\
\midrule

& \multicolumn{3}{c|}{\textbf{No Augmentation (\noaugb)}} & 36.6 & 60.2 & 64.1 & 41.5 & 68.5 & 49.9 & 28.0 & 49.8  \\
& \multicolumn{3}{c|}{\textbf{\zb{}}} & 35.6 & 59.2 & 68.6 & 50.1 & 63.6 & 49.6 & 29.7 & \;\textbf{50.9 } \\
\midrule
\multirow{1}{*}{(i)}  & Tanh Gate & \; \cmark & \; \xmark        & 35.0 & 57.8 & 55.8 & 49.9 & 71.5 & 51.6 & 27.9 & 49.9  \\
\midrule
\multirow{4}{*}{(ii)} &           & \; \multirow{4}{*}{5}    & \; 1      & 35.6 & 58.9 & 67.1 & 47.6 & 65.2 & 49.5 & 30.1 & 50.6 \\
                      & Num of        &                       & \; 10      & 35.3 & 59.4 & 62.1 & 46.3 & 64.6 & 51.4 & 29.4 & 49.8  \\
                      & Augs        &                       & \; 20$^\star$  & 34.7 & 58.7 & 60.3 & 46.5 & 61.6 & 50.1 & 28.4 & 48.6 \\
                      &       &                       & \; 30$^\star$  & 35.1 & 60.5 & 58.7 & 48.2 & 67.2 & 50.7 & 28.5 & 49.8  \\
\midrule
\multirow{2}{*}{(iii)} & Latent   & \;\multirow{2}{*}{64}   & \;32      & 34.9 & 58.8 & 64.3 & 44.5 & 67.9 & 51.2 & 28.6 & 50.0  \\
                       & size     &                       & \;128     & 34.7 & 57.6 & 63.1 & 47.8 & 69.8 & 50.4 & 28.1 & 50.2  \\
\midrule
\multirow{1}{*}{(iv)}  & Aug length      & \;\multirow{1}{*}{256} & \;512       & 35.3 & 58.8 & 58.6 & 52.5 & 68.9 & 50.3 & 28.8 & 50.5  \\
\midrule 
\multirow{2}{*}{(v)} & Training   & \;\multirow{2}{*}{\ours{}}    & \;No Aug+      & 37.6 & 58.4 & - & 43.3 & - & 50.7 & 28.0 & - \\
                     & mixture    &                             & \;\ours{}+     & 34.5 & 58.7 & - & 42.8 & - & 50.1 & 29.3 & - \\

\bottomrule

\end{tabular}
\caption{Ablation study. Each ablated setting should be compared with the first two rows, i.e., the original \textit{No Augmentation (\noaugb{})} setting and \textit{\zb{}}. The superscripted ``$\star$'' in ablated setting \textit{(ii)} indicates using the model variant with a frozen augmentation encoder. See descriptions of each setting in Section~\ref{sec:ablation}.}
\label{table:main_ablation}
\end{table*}

In this section, we continue investigating \textit{why \ours{} works} by conducting comprehensive ablation studies on different aspects of the model design. As shown in Table~\ref{table:main_ablation}, we consider the following five categories of ablated settings on \textit{\zb{}}~\footnote{We ablate on \zb{} instead of \zl{} mainly to reduce the computation overheads of a large amount of experiments.}:

\vspace{-5pt}
\paragraph{(i) Tanh gate.} We replace the gated cross-attention module with vanilla cross-attention in the ablated version. Specifically, we remove the two Tanh gates as shown in Figure~\ref{fig:architecture}. We find that \textbf{removing Tanh gate hurts the zero-shot performance}. Note that the Tanh gate is also the main difference between \textit{\ours{}} and \textit{FiD}~\citep{FiD}. 

\vspace{-5pt}
\paragraph{(ii) Number of augmentations.} We ablate on the number of augmentations. Note that for settings with 20 and 30 augmentations, in order to reduce the computation complexity, we propose another variant of \textit{\zb{}} where we encode augmentations with a separate frozen augmentation encoder. 
We find that increasing the number of augmentations from single to multiple (five) improves the performance. However, further increasing the number to 10 starts to hurt the performance, which again indicates that the noise starts to overwhelm the useful signals introduced by the retrieval. 
We also observe that the performance with 30 augmentations outperforms 20 augmentations, we hypothesis that this is due to inaccurate retrieval ranking that leads to some more informative documents being ranked lower. 
We show an example of this case in Figure~\ref{fig:example_inaccurate_ranking}.
Nevertheless, the fact that we are able to achieve positive gain with as many as 30 augmentations shows the \textbf{robustness of our model to very noisy augmentations}.

\vspace{-5pt}
\paragraph{(iii) Perceiver resampler latent size.} We ablate on the size of the latent query vector in the perceiver resampler. Note that here the latent size is different from the hidden state size of the backbone model. The trade-off of the size of the latent query vector is that, a larger latent size preserves more information from the original augmentation but also includes more noise. A larger latent size can also increase the computational complexity. We find that \ours{} is \textbf{relatively robust to the change of the latent size} and achieves the best performance with a latent size of 64.

\vspace{-5pt}
\paragraph{(iv) Per augmentation length.} We investigate the impact of different ways of constructing augmentations from the retrieved documents. Specifically, we increase the maximum length of each augmentation from 256 to 512 and fit two retrieved documents into one augmentation. We keep the number of augmentations the same as default, i.e., 5. We then compare this ablated setting with the 10-augmentation variant in (ii). We find that with the same set of retrieved documents, \textbf{augmenting the model with longer but fewer augmentations generally outperforms using a larger number of shorter augmentations}.

\vspace{-5pt}
\paragraph{(v) Training mixture.} We investigate the impact of adding new types of training tasks to the original training mixture. We dub the models trained with this new training mixture as \textit{\textbf{\noaug{}+}} and \textit{\textbf{\ours{}+}}. Specifically, apart from the eight multiple-choice QA tasks, we further include four more tasks: one closed-book QA task \textbf{WikiQA}~\citep{wikiqa}, one topic classification task \textbf{TREC}~\citep{trec}, one sentence completion task \textbf{COPA}~\citep{COPA}, and one sentiment task \textbf{Rotten Tomatoes} (RT)~\citep{rotten_tomatoes}~\footnote{We follow T0 to move tasks that are originally in the evaluation split, i.e. COPA and RT, into the training split in this ablated setting.}. We find that adding new types of tasks does not necessarily increase the performance. Although trained with only 8 tasks (v.s. 36 tasks) we are able to achieve state-of-the-art performance (Section~\ref{sec:compare_to_sota}), which shows that the \textbf{multiple-choice QA mixture is highly effective for generalizing to a wide range of held-out unseen tasks}.

\subsection{Analysis: computation overheads}
\label{sec:computation_overhead}
There are two main computation overheads compared with the fully-parametric counterpart, i.e., the \textit{No Aug} baseline. First, retrieving from a large-scale corpus can be time-consuming. As mentioned in Section~\ref{sec:c4_retrieval}, we apply document-level retrieval with BM25 and truncation on the query to reduce the retrieval time. We also perform the retrieval offline to avoid repeated time commitment. As a result, indexing 5\% of the C4 corpus takes 1 hour. Offline retrieval for the entire training and evaluation mixture takes 11 hours, which is approximately 0.28 seconds per instance. 
Furthermore, we measure the computation overhead on inference which is caused by the additional retrieved inputs as well as a small amount of newly introduced parameters (+4.6\%). The average computation overhead across all evaluation datasets during inference is around 4x compared with the \textit{No Aug} baseline. Notably, Table~\ref{table:main_result} shows that \zb{} achieves competitive performance with T0-3B while being 15x smaller in scale, indicating that the benefit of the retrieval augmentation overwhelms the computation overhead.

\section{Related Work}

\subsection{Semi-parametric models}

Semi-parametric models~\citep{virtual_kb, memoryover, DrQA,ORQA,REALM,multi_passage_bert,DPR,BERTserini,RAG,FiD}, which augmenting a parametric neural network with external knowledge bases or text corpora, have been widely applied to knowledge-intensive NLP tasks such as open-domain question answering. 
Recent advancements in semi-parametric \textit{language models}~\citep{knn-lm, spalm, retro, memory_aug_lm} have demonstrated improved language modeling performance with a relatively small language model and a retrieval system based on a large-scale corpus. 
Although the aforementioned semi-parametric language models have shown competitive performance on language modeling, 
compared with fully-parametric counterparts such as GPT-3~\citep{gpt3}, 
it is unclear whether the superiority generally holds on downstream tasks.
While concurrent work~\citep{meta_atlas} showed initial success in few-shot settings relying on Fusion-in-Decoder (FiD)~\citep{FiD} framework, this work focus on the more challenging \textbf{zero-shot} settings~\cite{t0, prompt_consistency, learning_inst_unlabel_zero_ta_gen}. Furthermore, instead of reusing FiD framework as in \cite{meta_atlas}, we show that our newly proposed fusion module is more effective than FiD due to the \textbf{gated mechanism}, which is inspired by Highway Networks~\citep{highway_networks, highway_transformer}, Gated Convolution~\cite{ lm_gated_cnn} and Vision-Language Fusion\citep{flamingo}.


\subsection{Massive multitask prompted training}
Based on the assumption that the reasonable zero-shot ability of large language models may come from implicit multitask learning during pretraining, recent studies~\citep{t0, FLAN, CrossFit, benchmarking1600} have demonstrated that explicitly training a language model on a mixture of diverse tasks can effectively improve its zero-shot performance on unseen tasks. 
In this work, we extend T0's multitask prompted training to a \textbf{semi-parametric} setting,
where we further augment the training and evaluation instances with retrieved documents. Notably, our work is distinguished from previous work ReCross~\citep{Lin2022UnsupervisedCG}, which uses upstream training data for augmentation, in twofold. First, we retrieve documents from a much larger task-agnostic corpus instead of clean upstream training instances. Second, in addition to directly concatenating the augmentation with the input just as FiD~\citep{FiD}, we further propose a novel \fusion{} module to handle retrieval noise.

\subsection{Fusion of retrieved augmentations}
\label{sec:related_work_fusion_of_augmentation}
In this work, the main challenge of designing the semi-parametric language model architecture is how to effectively leverage potentially noisy retrieved documents.
Existing methods on incorporating external texts fall in two categories, \textit{direct concatenation}~\citep{Lin2022UnsupervisedCG, gpt3, generated_knowledge_prompt, RAG, wang2022training} and \textit{cross-attention}~\citep{FiD, DoHA, retro}. 
However, we find that prior work lacks an explicit design for preventing the model from attending to noisy augmentations. 
Inspired by recent visual language models~\cite{flamingo,yu2022coca,li2022blip,jiang2022vima}, we find that we can actually borrow ideas from vision-language fusion for text-text fusion. We identify two key differences from Flamingo architecture: first, we use a much smaller encoder-decoder model that is jointly trained with the newly initialized layers instead of frozen layers. Second, instead of inserting the gated cross-attention module into a large frozen language model~\citep{chinchilla}, we add only one layer of gated cross-attention on top of the encoder to alleviate the need for extensive pre-training.

\section{Conclusion}
\label{sec:limitation}
In this work, for the first time, we show that semi-parametric language models have the potential to exhibit strong zero-shot task generalization ability by introducing \ours{}. 
Through extensive analysis and ablation study, we further demonstrate that the interplay of the proposed \fusion{}
 and the semi-parametric multitask training is essential towards \ours{}'s empirical success. Notably, our proposed \zl{} model outperforms T0-3B by 16\% across seven diverse evaluation tasks while being 3.8x smaller in scale.

\section{Limitation} 
In Section~\ref{sec:compare_to_sota}, we show that our training mixture with multiple-choice QA tasks, although small, is highly effective for multitask training. However, it is still unclear why multiple-choice QA tasks are particularly effective. Identifying the key factors towards positive or negative transfer from different tasks in the multitask training mixture would greatly help improve zero-shot task generalization. Future work includes investigating why certain mixtures are more effective than others and expanding the evaluation set to a wider range of tasks.
Computation overhead is another noticeable limitation of semi-parametric models which is discussed in detail in Section~\ref{sec:computation_overhead}.
Moreover, future work on developing more efficient and accurate retrieval methods for retrieving from large-scale task-agnostic corpus can definitely improve semi-parametric language models.

\section*{Acknowledgements}
We would like to express our gratitude to the anonymous reviewers for their insightful comments and suggestions. We would also like to thank our colleagues and fellow interns at Tencent AI Lab for their valuable internal discussions and feedback, as well as the students from Blender Lab at the University of Illinois Urbana-Champaign for their insightful feedback.

\bibliography{anthology,custom}
\bibliographystyle{acl_natbib}


\newpage

\appendix

\section{Qualitative analysis of the retrieved documents}
\label{sec:retrieval_examples}
\begin{figure*}[thb]
    \centering
    \includegraphics[width=\textwidth]{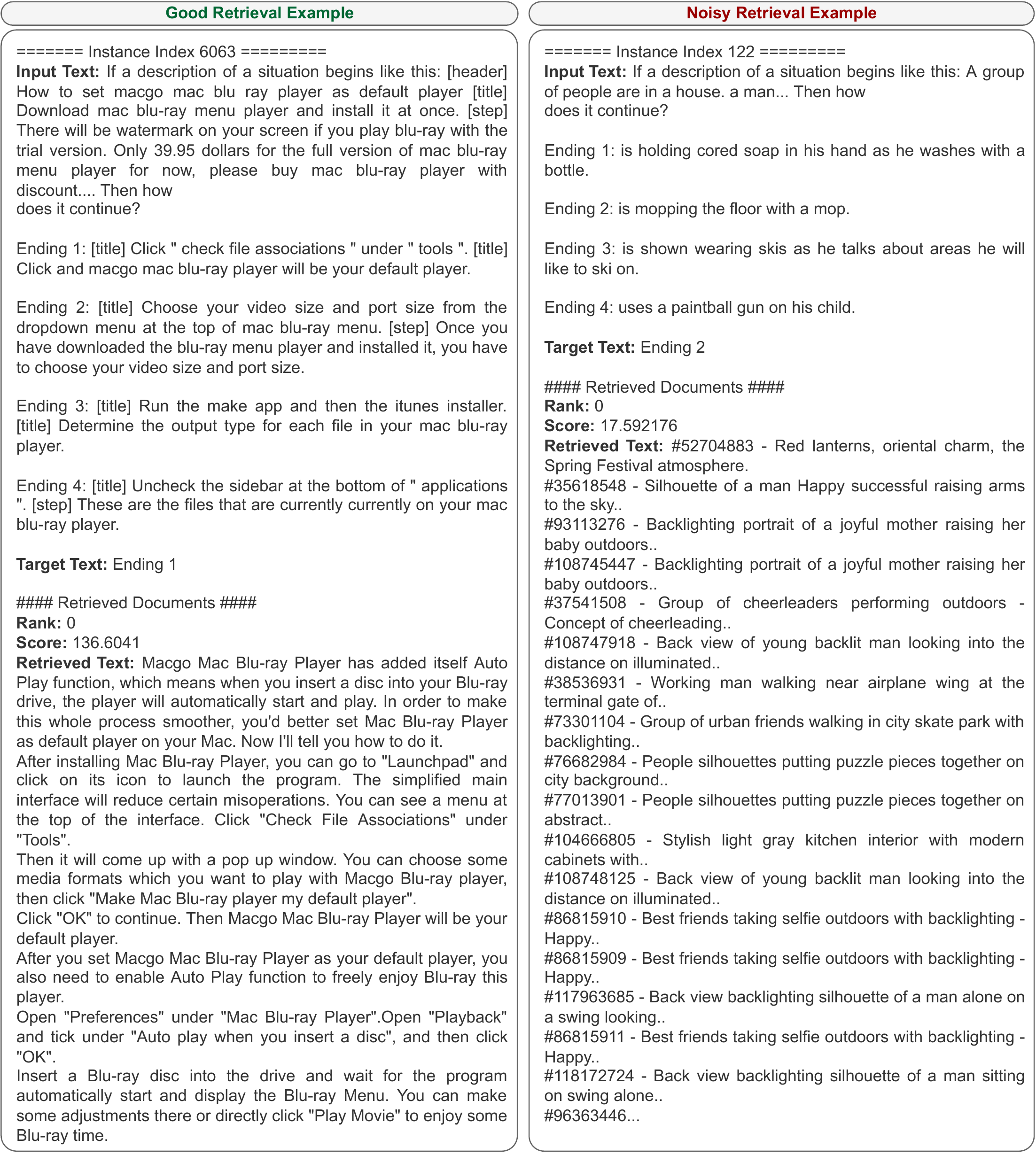}
    \caption{Example of retrieved documents on HellaSwag.}
    \label{fig:example_hellaswag}
\end{figure*}

\begin{figure*}[thb]
    \centering
    \includegraphics[width=\textwidth]{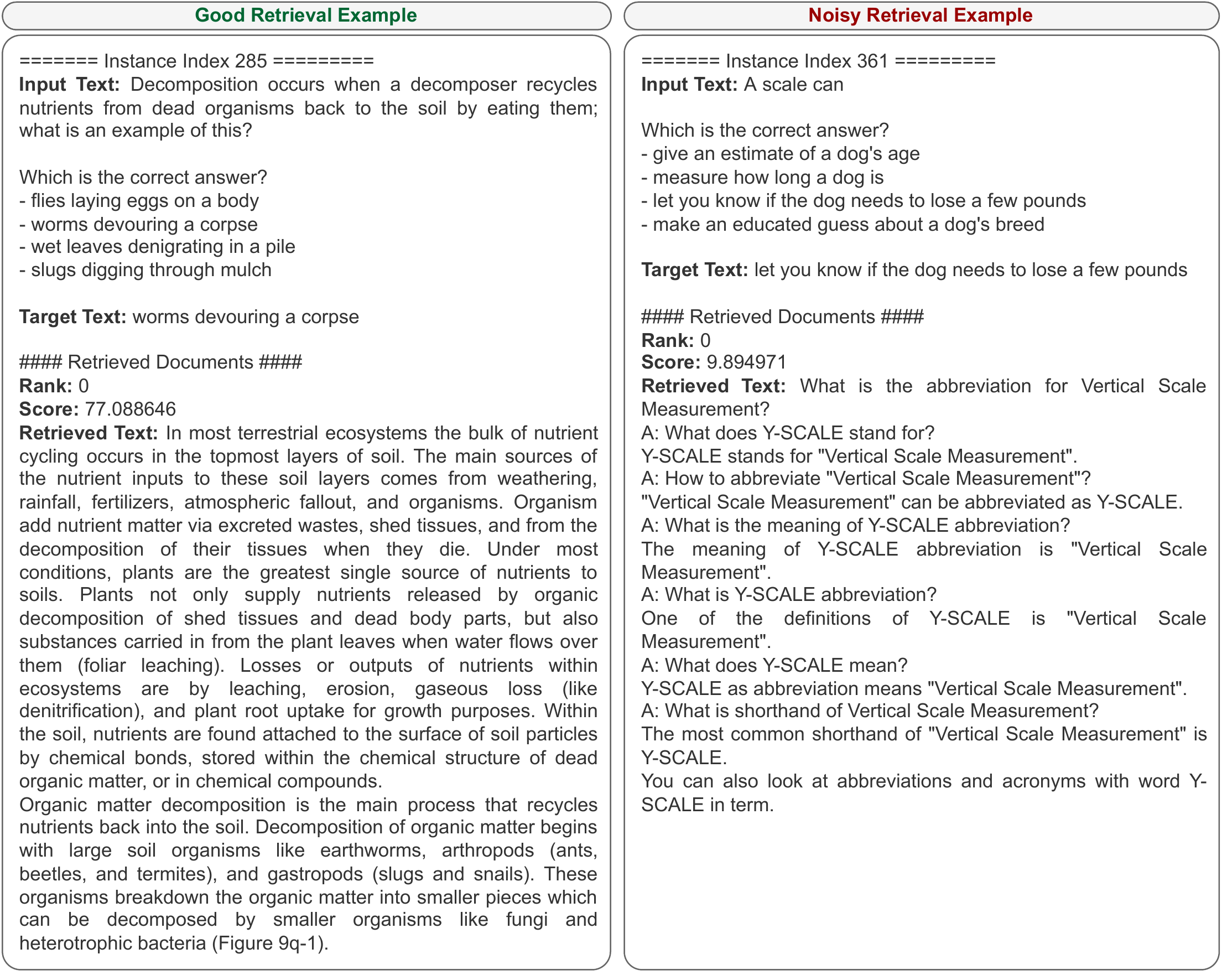}
    \caption{Example of retrieved documents on OpenbookQA.}
    \label{fig:example_obqa}
\end{figure*}

\begin{figure*}[thb]
    \centering
    \includegraphics[width=\textwidth]{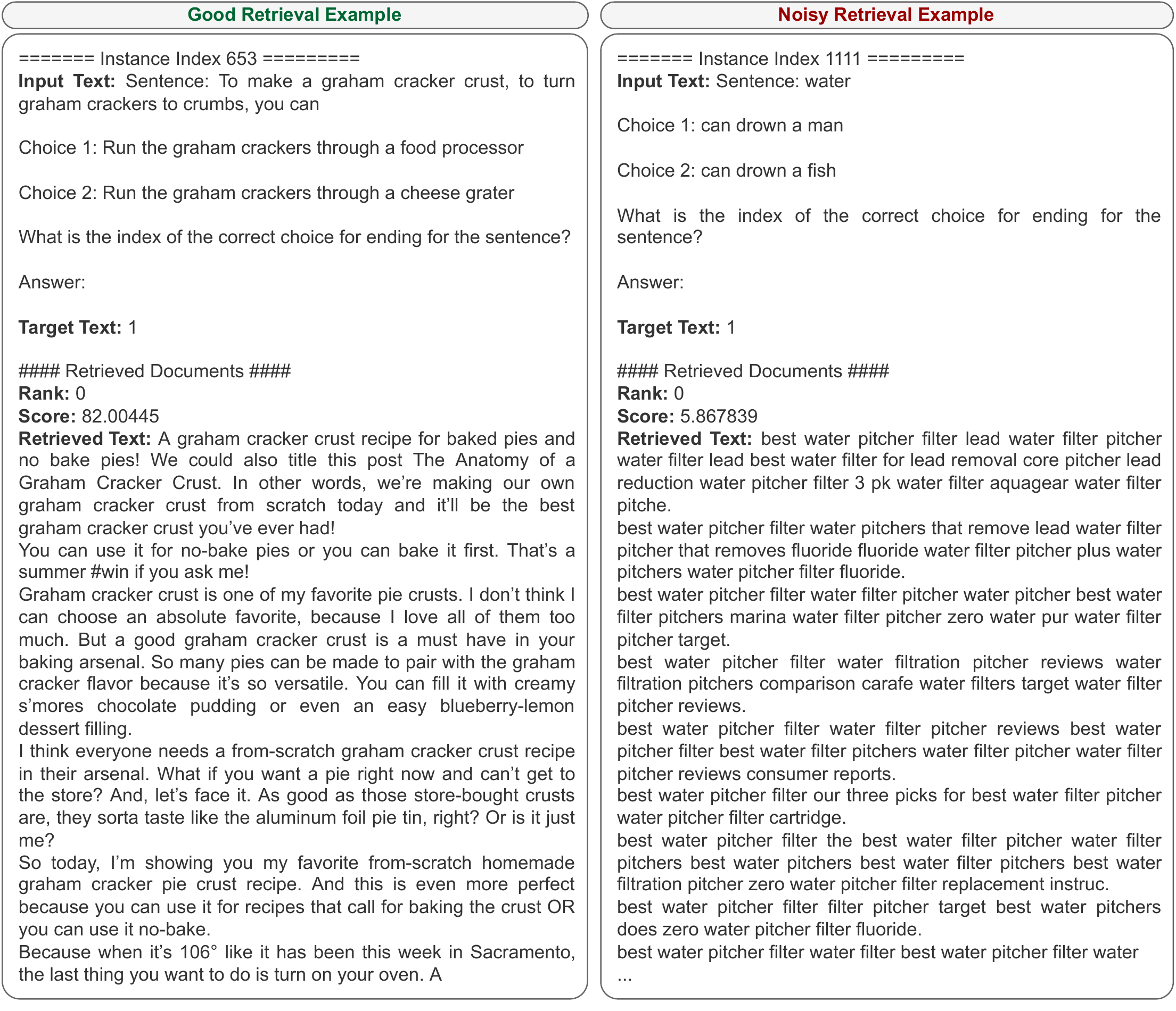}
    \caption{Example of retrieved documents on Piqa.}
    \label{fig:example_piqa}
\end{figure*}

\begin{figure*}[thb]
    \centering
    \includegraphics[width=\textwidth]{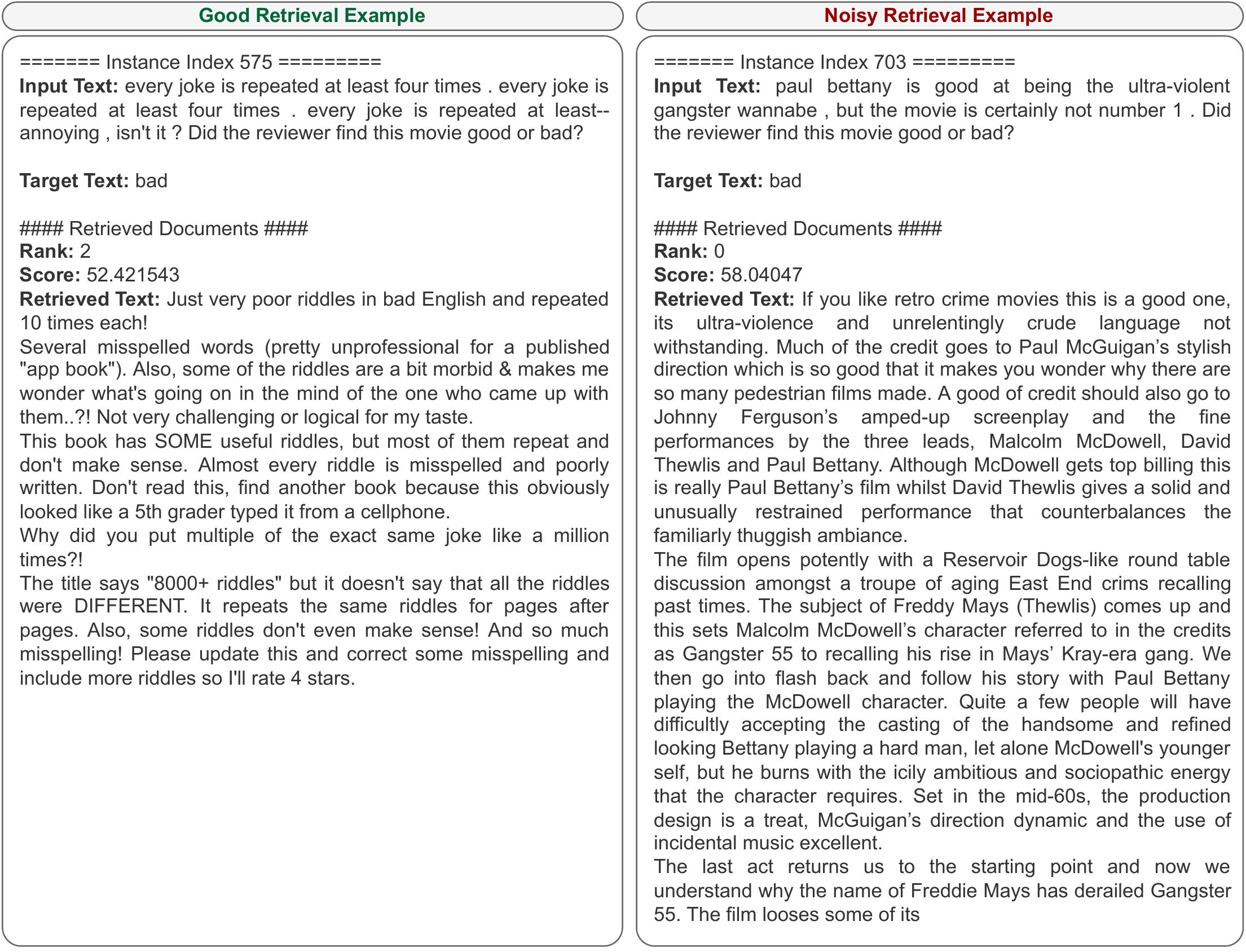}
    \caption{Example of retrieved documents on Rotten Tomatoes.}
    \label{fig:example_rt}
\end{figure*}

\begin{figure*}[thb]
    \centering
    \includegraphics[width=\textwidth]{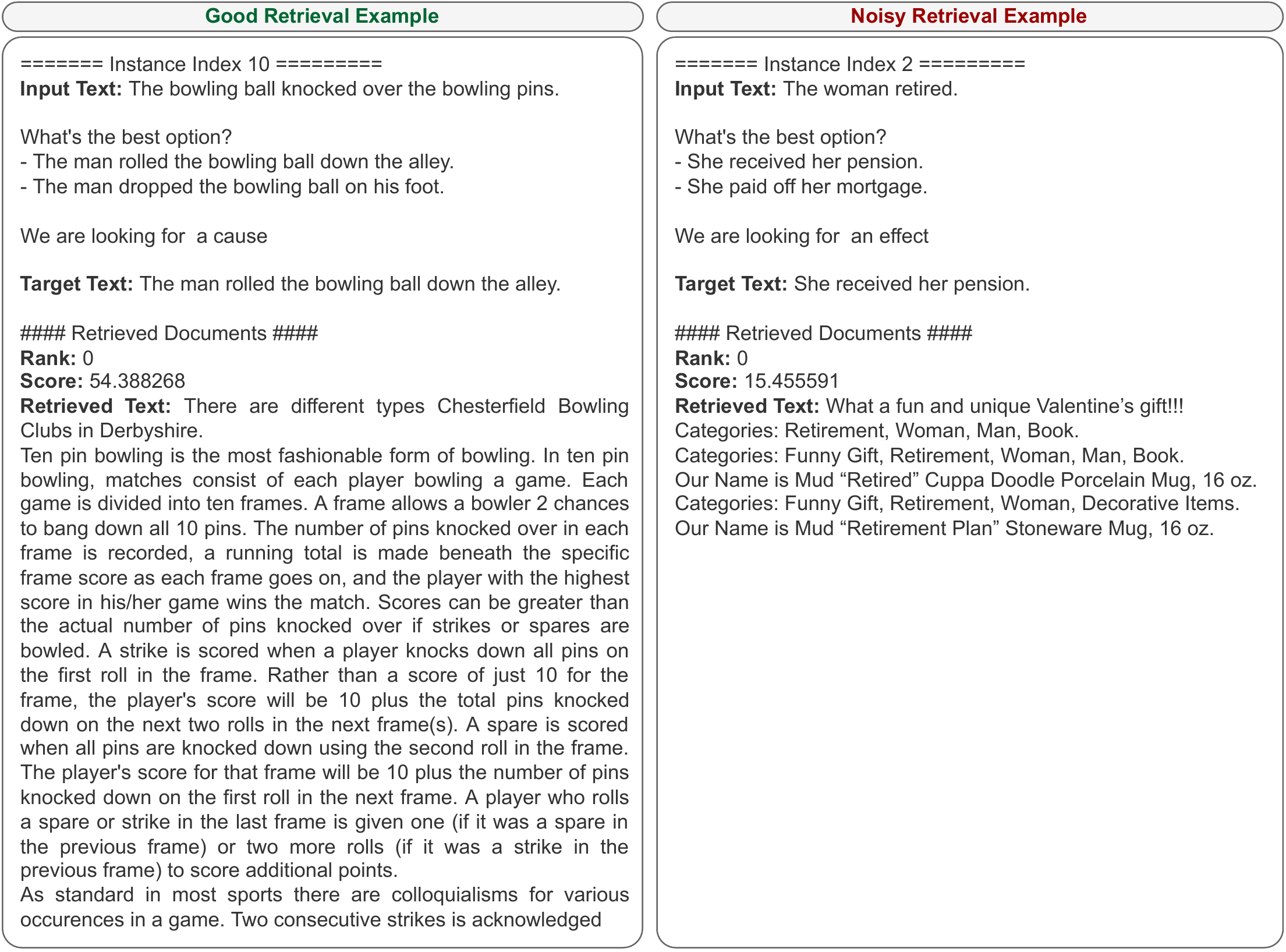}
    \caption{Example of retrieved documents on COPA.}
    \label{fig:example_copa}
\end{figure*}

\begin{figure*}[thb]
    \centering
    \includegraphics[width=\textwidth]{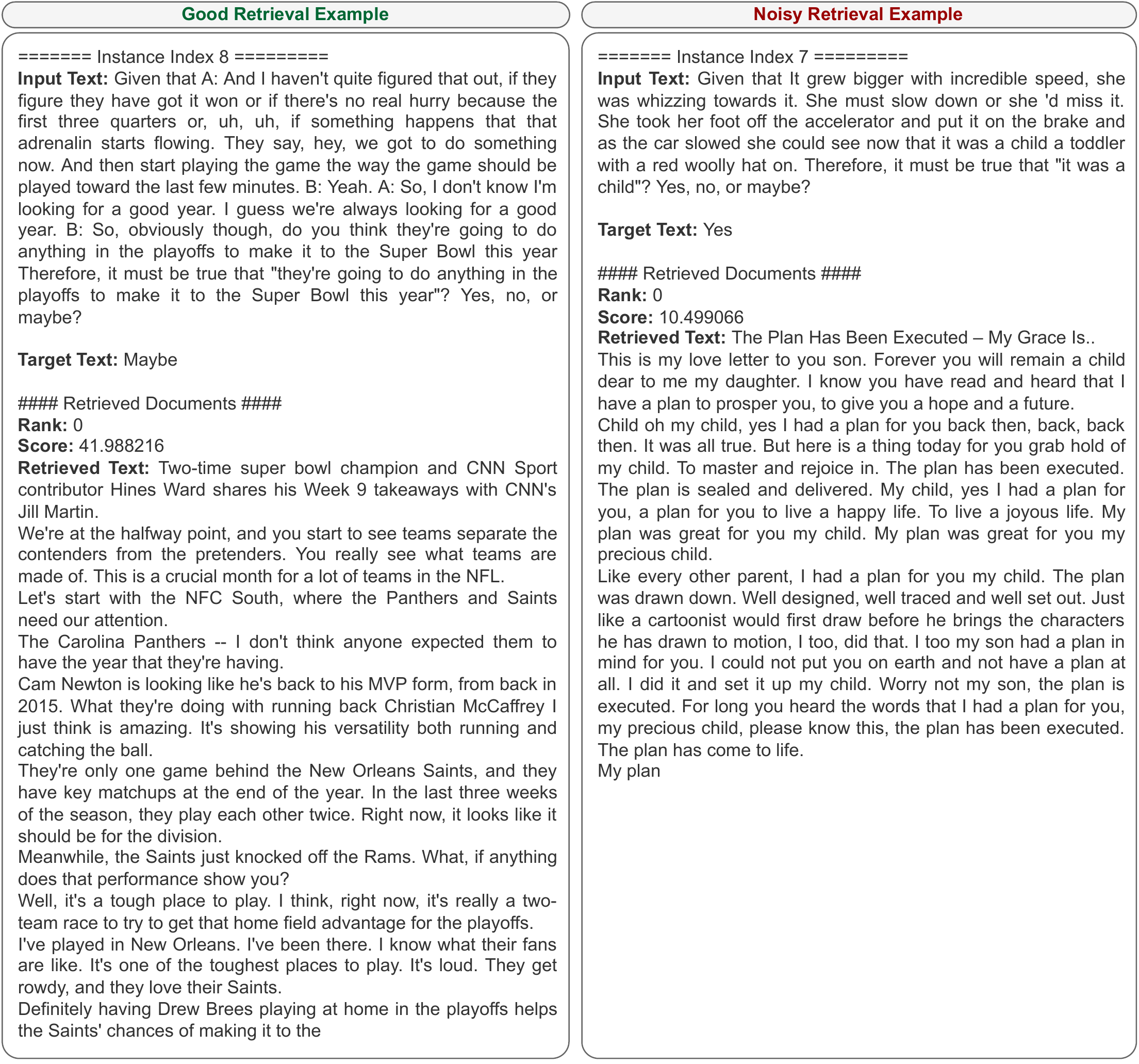}
    \caption{Example of retrieved documents on CB.}
    \label{fig:example_cb}
\end{figure*}

\begin{figure*}[thb]
    \centering
    \includegraphics[width=\textwidth]{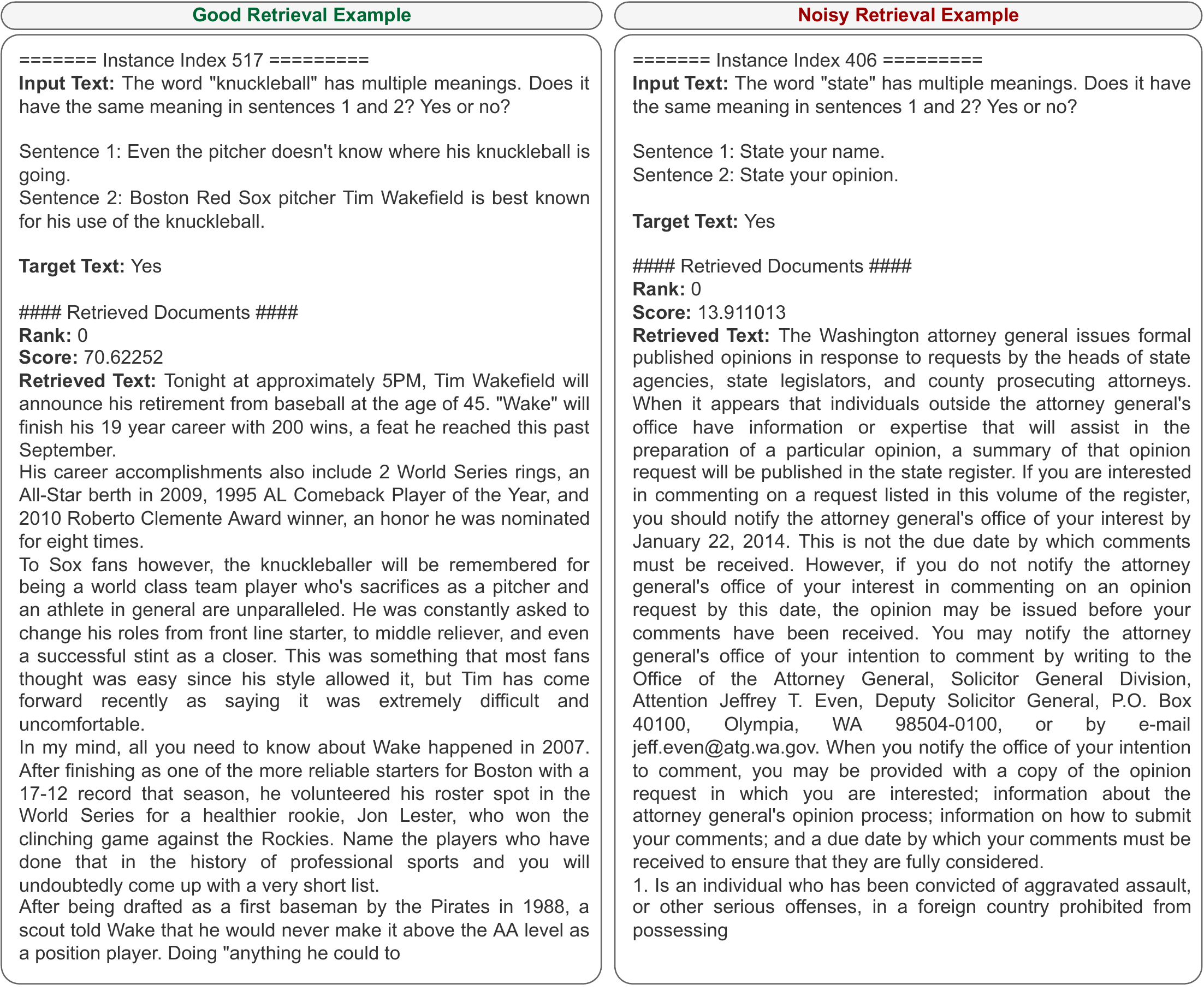}
    \caption{Example of retrieved documents on WiC.}
    \label{fig:example_wic}
\end{figure*}

\begin{figure*}[thb]
    \centering
    \includegraphics[width=\textwidth]{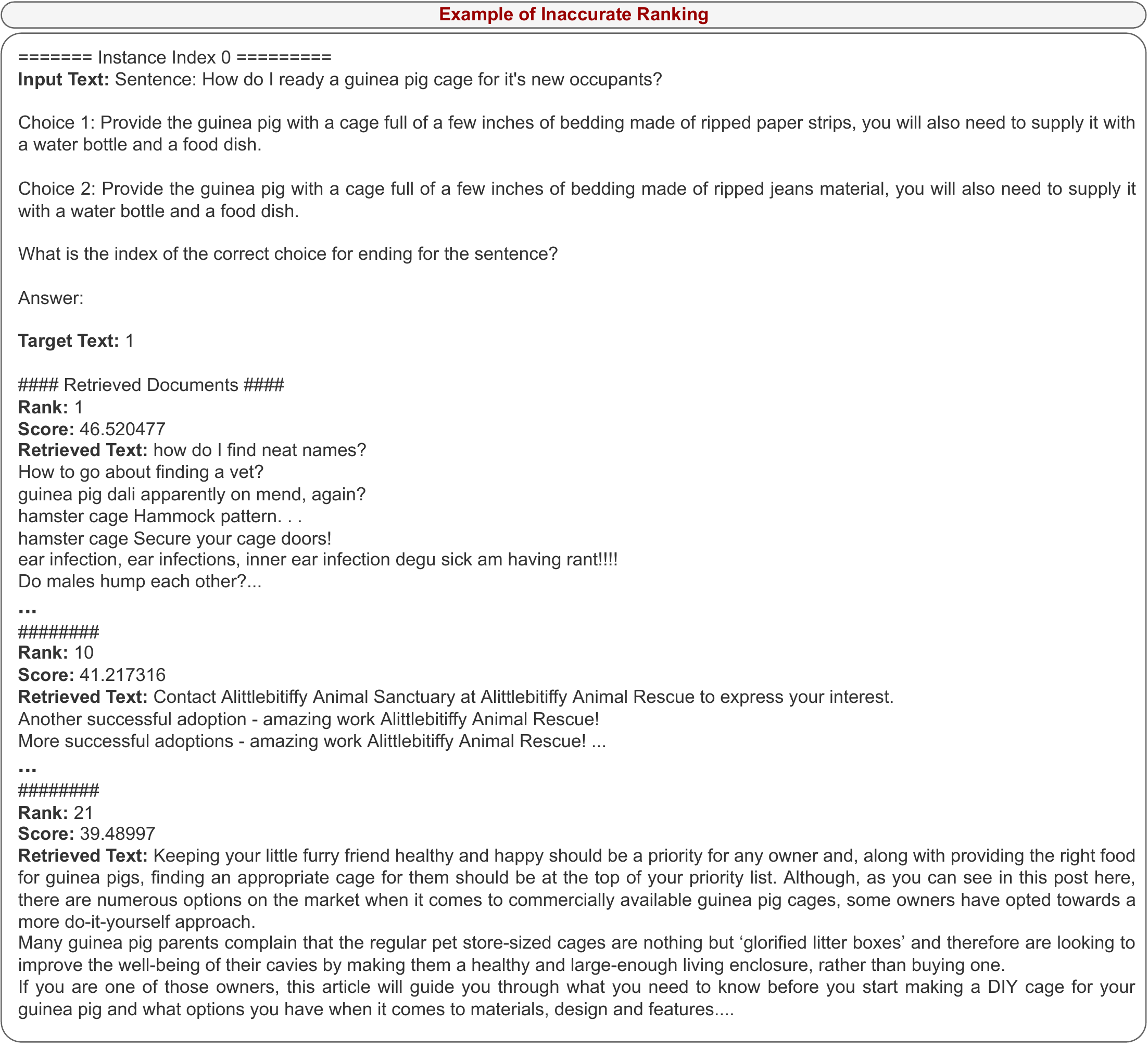}
    \caption{Example of the inaccurate ranking of the retrieval. Here we show the ranked retrieved documents for instance 0 in Piqa. We can see that the 21th ranked document is more correlated than many of the higher ranked ones, such as rank 1 and rank 10.}
    \label{fig:example_inaccurate_ranking}
\end{figure*}

Here we visualize one good and one noisy example of the retrieved documents for each evaluation task. A full list of examples for each training and evaluation task can be found in the supplementary material under the ``visualization'' folder. As shown in Figure~\ref{fig:example_hellaswag}, \ref{fig:example_obqa}, \ref{fig:example_piqa}, \ref{fig:example_rt}, \ref{fig:example_cb}, \ref{fig:example_copa}, and \ref{fig:example_wic}, the retrieved augmentations can contain highly correlated information that can be directly helpful for solving a certain task, however, they can also be very noisy. As mentioned in Section~\ref{sec:c4_retrieval}, the retrieved documents can also be inaccurately ranked, for example in Figure~\ref{fig:example_inaccurate_ranking}, we show that the 21th ranked retrieval result can contain more correlated information than the top ranked ones. Furthermore, as shown in the noisy example of Figure~\ref{fig:example_rt}, for some tasks such as sentiment analysis, even though the retrieved document is highly correlated with the input text, i.e., with a high BM25 score, the content can steer the prediction into a wrong direction. These observations motivate us to propose the augmentation fusion module with a gated mechanism.

\section{Implementation details}
\label{app:implementation_details}
We use T5-base and T5-large as backbone model for \zb{} and \zl{}, respectively. We follow \cite{flamingo} to implement the perceiver resampler and the gated cross-attention. For both \zb{} and \zl{}, unless otherwise specified, we use one layer of gated cross-attention and one layer of perceiver resampler with a latent size of 64.  A comprehensive ablation study on the impact of different aspects of our model design such as the Tanh Gate can be found in Section~\ref{sec:ablation}. 
All models are trained on the same training mixture as mentioned in Section~\ref{sec:datasets} for ten epochs with a batch size of 32 and a learning rate of 1e-4. We report results from the checkpoint that achieved the best overall performance across all tasks. All experiments are done on eight NVIDIA-V100 32GB GPUs.

\section{Full list of tasks and templates}
\label{app:templates}
Following T0~\citep{t0}, we use tasks from Hugginface Datasets~\citep{hfdatasets} and templates from PromptSource~\citep{promptsource} marked as ``original task" and with ``choices\_in\_prompt". Specifically, for tasks in the training mixture, we randomly sample two templates per task for semi-parametric multitask prompted training. For tasks in the held-out evaluation mixture, we use all available templates. Table~\ref{table:templates_part1}, and \ref{table:templates_part2} shows the full list of templates we used for each task during multitask training and zero-shot evaluation. 
\begin{table*}[thb]
\small
\centering
\setlength\tabcolsep{3pt}
\begin{tabular}{c | c | c }
\toprule
\multirow{1}{*}{\textbf{Mixture}} &
\multirow{1}{*}{\textbf{Task}} &
\multirow{1}{*}{\textbf{Template Name}} \\
\midrule
\multirow{16}{*}{\textbf{Semi-T0 Training}}  & \multirow{2}{*}{cos\_e/v1.11} & question\_option\_description\_text \\
                                    &                                & description\_question\_option\_id \\
                                    \cmidrule{2-3}
                                    & \multirow{2}{*}{cosmos\_qa}    & context\_description\_question\_answer\_id \\
                                    &                                & description\_context\_question\_answer\_text \\
                                    \cmidrule{2-3}
                                    & \multirow{2}{*}{dream}         & baseline \\
                                    &                                & read\_the\_following\_conversation\_and\_answer\_the\_question\\
                                    \cmidrule{2-3}
                                    & \multirow{2}{*}{qasc}          & qa\_with\_separated\_facts\_1 \\
                                    &                                & qa\_with\_separated\_facts\_4 \\
                                    \cmidrule{2-3}
                                    & \multirow{2}{*}{quartz}        & answer\_question\_below \\
                                    &                                & read\_passage\_below\_choose\\
                                    \cmidrule{2-3}
                                    & \multirow{2}{*}{sciq}          & Multiple Choice \\
                                    &                                & Multiple Choice Question First \\
                                    \cmidrule{2-3}
                                    & \multirow{2}{*}{social\_i\_qa} & Show choices and generate answer \\
                                    &                                & Show choices and generate index \\
                                    \cmidrule{2-3}
                                    & \multirow{2}{*}{wiqa}          & effect\_with\_string\_answer \\
                                    &                                & effect\_with\_label\_answer \\
\midrule
\multirow{8}{*}{\textbf{Semi-T0+ Training}}  & \multirow{2}{*}{wiki\_qa}         & Decide\_good\_answer \\
                                    &                                  & found\_on\_google \\
                                   \cmidrule{2-3}
                                    & \multirow{2}{*}{trec}            & what\_category\_best\_describe \\
                                    &                                  & trec1 \\
                                   \cmidrule{2-3}
                                    & \multirow{2}{*}{super\_glue/copa} & more likely \\
                                    &                                  & best\_option \\
                                   \cmidrule{2-3}
                                    & \multirow{2}{*}{rotten\_tomatoes} & Sentiment with choices \\
                                    &                                  & Reviewer Opinion bad good choices \\
\bottomrule

\end{tabular}

\caption{PromptSource template names used for each task (Part1).}
\label{table:templates_part1}
\end{table*}

\begin{table*}[thb]
\small
\centering
\setlength\tabcolsep{3pt}

\begin{tabular}{c | c | c }
\toprule
\multirow{1}{*}{\textbf{Mixture}} &
\multirow{1}{*}{\textbf{Task}} &
\multirow{1}{*}{\textbf{Template Name}} \\
\midrule

\multirow{50}{*}{\textbf{Semi-T0 Evaluation}}  & \multirow{2}{*}{openbookqa/main}  & choose\_an\_answer\_with\_options \\
                                                                        &&  which\_correct \\
                                                                        &&  pick\_using\_id \\
                                                                        && choices \\
                                                                        && only\_options \\
                                                                        && which\_correct\_inverse \\
                                                                        && pick\_answer\_with\_options\\
                                    \cmidrule{2-3}
                                    & \multirow{2}{*}{piqa}            &   what\_is\_the\_correct\_ending \\
                                                                       &&  pick\_correct\_choice\_with\_choice\_given\_before\_goal \\
                                                                       &&  pick\_correct\_choice\_index \\
                                                                       &&  finish\_sentence\_with\_correct\_choice \\
                                                                       &&  choose the most appropriate solution \\
                                   \cmidrule{2-3}
                                    & \multirow{2}{*}{rotten\_tomatoes} &  Reviewer Opinion bad good choices\\
                                                                       && Sentiment with choices \\
                                   \cmidrule{2-3}
                                    & \multirow{2}{*}{super\_glue/cb}    & can we infer \\
                                                                        && based on the previous passage \\
                                                                        && claim true/false/inconclusive \\
                                                                        && does it follow that \\
                                                                        && justified in saying \\
                                                                        && always/sometimes/never \\
                                                                        && GPT-3 style \\
                                                                        && consider always/sometimes/never \\
                                                                        && guaranteed true \\
                                                                        && must be true \\
                                                                        && guaranteed/possible/impossible \\
                                                                        && does this imply \\
                                                                        && MNLI crowdsource \\
                                                                        && should assume \\
                                                                        && take the following as truth \\
                                   \cmidrule{2-3}
                                    & \multirow{2}{*}{super\_glue/copa} & exercise \\
                                                                        && …What could happen next, C1 or C2?\\
                                                                        && i\_am\_hesitating\\
                                                                        && plausible\_alternatives\\
                                                                        && C1 or C2? premise, so/because…\\
                                                                        && …As a result, C1 or C2?\\
                                                                        && best\_option\\
                                                                        && …which may be caused by\\
                                                                        && more likely\\
                                                                        && cause\_effect\\
                                                                        && …why? C1 or C2\\
                                                                        && choose\\
                                   \cmidrule{2-3}
                                    & \multirow{2}{*}{super\_glue/wic}  & question-context-meaning-with-label\\
                                                                        && grammar\_homework\\
                                                                        && affirmation\_true\_or\_false\\
                                                                        && same\_sense\\
                                                                        && GPT-3-prompt-with-label\\
                                                                        && polysemous\\
                                   \cmidrule{2-3}
                                    & \multirow{2}{*}{hellaswag}       & complete\_first\_then \\
                                                                       &&  Randomized prompts template\\
                                                                       && Predict ending with hint\\
                                                                       && if\_begins\_how\_continues\\
\bottomrule

\end{tabular}

\caption{PromptSource template names used for each task (Part2).}
\label{table:templates_part2}
\end{table*}

\section{Retrieval query key for each task}
\label{app:query_key}
\begin{table*}[thb]
\small
\centering
\setlength\tabcolsep{3pt}
\begin{tabular}{ c | c }
\toprule
\multirow{1}{*}{\textbf{Task}} &
\multirow{1}{*}{\textbf{Query Key}} \\
\midrule
 cos\_e/v1.11 & question \\
 cosmos\_qa & question \\
 dream & question \\
 qasc & question \\
 quartz & question \\
 sciq & question \\
 social\_i\_qa & context \\
 wiqa & question\_stem \\
 openbookqa/main & question\_stem \\
 piqa & goal \\
 rotten\_tomatoes & text \\
 super\_glue/cb & hypothesis \\
 super\_glue/copa & premise \\
 super\_glue/wic & sentence1 \\
 hellaswag & ctx \\
 wiki\_qa & question \\
 trec & text \\
\bottomrule

\end{tabular}
\caption{Retrieval query key used for each task.}
\vspace{-5pt}
\label{table:query_key}
\end{table*}
In order to retrieve most relevant documents for each instance, we specify a certain field for each dataset which will be served as the query to the retrieval system. For example, for most multiple-choice QA tasks, we use the ``question'' string as our query. Table~\ref{table:query_key} shows a full list of field names we use as retrieval query keys for each dataset. Note that the field name shown in the table is what appears to be in the corresponding Huggingface Dataset~\citep{hfdatasets}.


\section{Broader impact}
One major benefit of developing powerful semi-parametric language models is that we can reduce the negative environmental impact from training huge parametric models. 
However, since the backbone language model is pretrained on massive internet-scale text data, there might be unexpected output that can have potential negative impact on the society, such as bias against people of a certain gender, race or sexuality. We are fully aware of the risks of potential misuses and will actively work with the community to improve the responsibility of large NLP models.

\end{document}